\pdfoutput=1

\documentclass[11pt]{article}

\usepackage[]{rs2022}

\usepackage{times}
\usepackage{latexsym}

\usepackage[T1]{fontenc}

\usepackage[utf8]{inputenc}

\usepackage{microtype}
\usepackage{booktabs}
\usepackage{arydshln}
\usepackage{graphicx}
\usepackage{amsmath}
\usepackage{subcaption}
\usepackage{soul}

\title{A Systematic Evaluation of Response Selection for Open Domain Dialogue}

\author{
Behnam Hedayatnia, Di Jin, Yang Liu, Dilek Hakkani-Tur\\
Amazon Alexa AI\\
\texttt{\{behnam,djinamzn,yangliud,hakkanit\}@amazon.com}
}

\begin{document}
\maketitle
\begin{abstract}

Recent progress on neural approaches for language processing has triggered a resurgence of interest on building intelligent open-domain chatbots. However, even the state-of-the-art neural chatbots cannot produce satisfying responses for every turn in a dialog. A practical solution is to generate  multiple response candidates for the same context,
and then perform response ranking/selection to determine which candidate is the best. Previous work in response selection typically trains response rankers using synthetic data that is formed from existing dialogs by using a ground truth response as the single appropriate response and constructing inappropriate responses via random selection or using adversarial methods. In this work, we curated a dataset where responses from multiple response generators produced for the same dialog context are manually annotated as appropriate (positive) and inappropriate (negative). We argue that such training data better matches the actual use case examples, enabling the models to learn to rank responses effectively. 
With this new dataset, we conduct a systematic evaluation of state-of-the-art methods for response selection, and demonstrate that both strategies of using multiple positive candidates and using manually verified hard negative candidates can bring in significant performance improvement in comparison to using the adversarial training data, e.g., increase of 3\% and 13\% in Recall@1 score, respectively. 

\end{abstract}

\section{Introduction}

Building an open-domain dialog system to interact with users on a variety of topics can involve building multiple response generators (RG) with different functions~\citep{paranjape2020neural}. These RGs can be a mixture of generative, retrieval and template based methods. 
A response selector is then built to re-rank response candidates produced by different applicable RGs to determine the best response for a given turn. 
These response selectors are based on either rule-based or model-based architectures~\citep{papaioannou2017alana, serban2017deep, zhou2020design, see-manning-2021-understanding}. Rule-based systems typically consist of manually-designed logic to rank hypotheses, whereas model-based approaches can either be conventional machine learning models or recent neural models that learn to rank candidates. As the number of RGs grows, a rule-based system can become cumbersome to maintain, whereas model-based methods can simplify the selection process as well as achieve better performance. 

Latest work in model-based response selectors involves leveraging pretrained transformer models such as BERT~\cite{devlin2018bert} and DialoGPT~\cite{zhang2019dialogpt}. 
These selection models are often trained using existing dialog datasets that typically contain ground truth responses.   
Thus a focus of past response selection work is on the construction of inappropriate/negative responses, using methods such as random selection, utterance manipulation or leveraging user feedback~\citep{whang2020effective, han-etal-2021-fine,whang2021response, gu2020speaker, xu2020learning, zhang2021structural, gao2020dialogue, see2021understanding, gupta2021synthesizing, li-etal-2019-sampling}. 
However, such synthesized datasets for response selection have the following known drawbacks. First of all, their claimed incorrect responses are not  verified if they are actually incorrect. Second, these negative responses are easy to differentiate from positive ones since it is very likely that they will be on different topics from the context. Therefore, models trained on such easy negative responses will not be able to generalize to real-world settings, where multiple responses are generated given the same dialog context and many of them are strong candidates.

To resolve the aformentioned issues, we construct a new dataset (named RSD) for response selection by showing human annotators multiple response candidates produced by different RGs for a given turn and dialog context, and asking them to annotate all responses that are appropriate for that specific dialog context. 
We leverage RSD to conduct a systematic evaluation of state-of-the-art methods for response selection, including existing trained models,  DialogRPT \cite{gao2020dialogue} and BERT-FP \cite{han-etal-2021-fine}, and a BERT based ranker that we trained. 
Our experimental results show the following findings: (1) Models trained on RSD significantly outperform those trained on existing datasets, e.g., Reddit and Ubuntu, showing the benefit of bringing in human annotated data for this task; (2) Using manually verified hard negatives greatly outperforms using adversarial negatives; (3) Training on multiple positive candidates improves performance in comparison to a single positive candidate.
Though these findings are most expected, this is the first empirical study that clearly shows that constructing a more realistic dataset benefits strongly over generating synthetic examples for response selection, and we hope such results can guide future research in this direction and deployment of open domain dialog systems.

\vspace{-0.05in}
\section{Related Work}

Previous work in response selection has been conducted in different domains, such as chat-logs~\cite{lowe2015ubuntu}, e-commerce~\cite{zhang2018dua}, and open-domain dialog~\citep{wu2017sequential, zhang2018personalizing, smith2020can, see2021understanding}. Our work focuses on open-domain dialog, where current systems typically consist of multiple response generators, each of which is designed to deal with a certain domain. For example, in the Alexa Prize challenge~\citep{ram2018conversational, gabriel2020further}, most of the participating socialbots built by university teams consist of a variety of responders that are based on retrieval-based methods, template-based methods, or generative models~\citep{konrad2021alquist,saha2021proto, paranjape2020neural,ram2018conversational}. In order to select the final response to present to users, both rule-based or model-based ranking  models have been proposed \cite{ram2018conversational, papaioannou2017alana, serban2017deep, zhou2020design, see-manning-2021-understanding, shalyminov2018neural}. This approach is also common in other real-world systems such as XiaoIce that employs a manually-designed set of features to rank hypotheses~\cite{zhou2020design}.

For training response selection models, typically human-human dialogs are used, where positive examples are the ground truth responses and negative responses are often randomly selected or synthetically created since there are no labeled negative responses. \citet{han-etal-2021-fine} randomly selected responses from other dialogs or within the same dialog session. \citet{whang2021response} corrupted utterances by inserting, substituting and deleting random tokens. \citet{xu2020learning} masked and shuffled utterances within a dialog. 
\citet{li-etal-2019-sampling} selected negative responses from a batch based on their similarity scores from the positive response score.
\citet{gupta2021synthesizing} used automatic methods such as replacing random tokens in a positive examples using a Mask-and-fill approach to create adversarial negative examples.

However, these sampling strategies do not ensure the selected negative responses are hard examples. In this work, rather than relying on approximation for negative responses, we perform turn level annotation of multiple response candidates for response appropriateness for a given dialog context.
\citet{see2021understanding, gao2020dialogue} did construct hard negative examples by annotating responses from a single generative model for appropriateness; however, our work contains responses from a mixture of various RG methods.

On the other hand, open-domain dialogs can have multiple  appropriate responses for a given dialog context. Previous work has augmented dialog datasets with multiple positive examples~\citep{mizukami2015adaptive, khayrallah2020smrter, gupta2019investigating, sai2020improving,zhang2020task}. 
Within open-domain dialogs, \citet{gupta2019investigating,sai2020improving} augmented the DailyDialog dataset \cite{li2017dailydialog} with multiple positive human written responses. In contrast, our dataset has multiple positive responses generated from models, which reduces the cost of human annotation significantly. The closest work to ours is \cite{sai2020improving} that constructed negative examples by asking  annotators to copy information from the dialog context. We do not restrict the definition of negative examples to be copying information from the dialog context, since incorrect responses in open-domain dialog can have different issues, e.g., off-topic, contradicting or repetitive responses. 

\begin{figure*}
 	\centering
 		\includegraphics[width=1.0\textwidth]{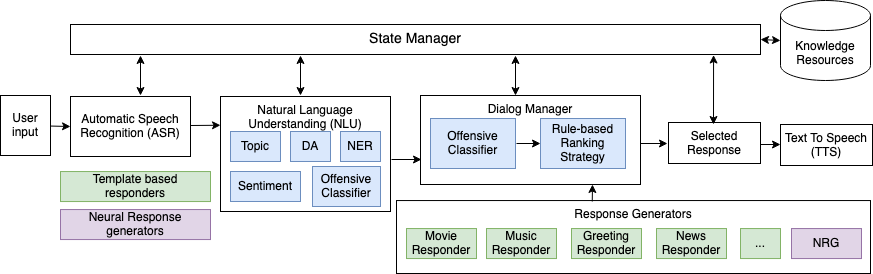}
 	\caption{Architecture of our Open Domain Dialog System. NER = Named Entity Recognition, DA = Dialog Act}
	\label{fig:dialog_system}
\end{figure*}

\section{Datasets}

As described earlier, most previous work in response selection has constructed test sets that typically contain only one positive candidate and one or more synthetically created negative candidates. However, such negative responses may be easy for a model to detect. Additionally, in real-world open-domain dialogs there can be more than one positive response per turn. 
Therefore, in this work we constructed a more realistic dataset consisting of annotations for real response candidates. Our dataset consists of spoken interactions between a dialog system and real users.

\subsection{Open Domain Dialog System}
\label{sec:dialog_system}

We first describe the open-domain dialog system used for data collection. The architecture of our dialog system is shown in Figure~\ref{fig:dialog_system}.
Every user utterance in the dialog is sent into an ASR system whose output goes through a series of NLU modules that classifies topics, dialog acts,  sentiment, extracts entities, and detects if user utterance is offensive. Our system then calls multiple response generators for the given dialog context and logs all the generated response candidates within the State Manager. The response presented to the user is selected by a rule-based ranker and then sent to the TTS module. 

For popular topics in open domain dialogs, such as movies, music, recent news, we developed template-based response generators (highlighted in green in Figure~\ref{fig:dialog_system}) for the given dialog state. An example state and response for the movie domain is: when the user turn mentions a movie name (based on the NER result), we respond with information about the actor, the rating, or the plot of this certain movie. In addition to topic-specific template-based RGs, our system includes other template-based RGs for different dialog contexts, such as, greetings, topic switches, etc.

For every user turn, we also apply a neural network-based response generation (NRG) model to produce a response, highlighted in purple in Figure~\ref{fig:dialog_system}. 
Our~\textit{NRG Responder} is a GPT2-XL~\cite{radford2019language} based model trained on real user conversation data described in Section~\ref{sec:data}. We discuss its training details in Appendix~\ref{nrg_training_details}. 

The rule-based ranker uses predefined logic and the topic extracted from the user utterance to select domain specific template-based responders. If a template-based responder is not available it will use the NRG response as a fall back. Our system has just a few template-based RGs, and uses NRG responses for almost half of all turns.

\begin{table*}[t]
\centering
\small
\resizebox{0.99\textwidth}{!}{
\begin{tabular}{c c c c c c}
\toprule
     Data Split & \# Dialogs & \# positive responses & \# negative responses & Avg. \# responses at each turn & \# Turns with no positive responses\\ \midrule
    
     RSD Train & 1,501 & 17,778 & 78,273 & 5.67 & 8,871\\
     RSD Test & 142 & 2,995 & 6,298 & 5.36 & 309\\ \bottomrule
 \hline
\end{tabular}}
\caption{\label{dataset_statistics} Dataset Statistics. For our experiments, we conduct 5 fold cross validation on all our training datasets and therefore do not have a dedicated development set. }
\end{table*}

\subsection{Response Selection Data (RSD)}
\label{sec:data}
We deploy the dialog system described above  within the Alexa Prize Socialbot framework~\cite{ram2018conversational} to interact with real users.
A user initiates an interaction with our dialog system and consents to have their data being collected. These interactions end when the user requests to stop the conversation. At the end of each interaction, users are asked to leave a rating in the range of 1 to 5. We denote this dataset as real user interactions (RUI)\footnote{All interactions are in English.}. Our data consists of approximately 100k interactions and 2.5 million turns. For each user turn in RUI, we produced additional response candidates using variants of our \textit{NRG Responder} to supplement the logged responses. These may be appropriate responses, or hard negative examples. The NRG variants we used include the following (Further model training details are in Appendix~\ref{nrg_training_details}).
\vspace{-0.05in}
 
\begin{itemize}
    \itemsep -0.75ex
    \item 
    A GPT2-medium version of our~\textit{NRG Responder}.
    
    \item A GPT2-XL~\textit{NRG Responder} grounded on knowledge. When there is an entity in the user turn, we search Wikipedia to find the article related to the entity, and perform knowledge selection and knowledge-grounded response generation. 
    
    \item A GPT2-medium~\textit{NRG Responder} grounded on dialog acts (DA)~\cite{hedayatnia2020policy}.

    \item A GPT2-XL based sentiment controlled ~\textit{NRG Responder}. When the user's utterance shows some negative sentiment (e.g., when a person says ``I'm depressed''), the NRG model generates a response conditioned on this emotion. 
    
\end{itemize}

We worked with internal human annotators to set up an annotation pipeline. These internal annotators are not experts in the dialog domain; however, we worked closely with them to ensure they have a clear understanding of the task provided to them. In our annotation pipeline, for each turn in a dialog, we showed internal human annotators all the available responses produced by the template based generators and various NRG models\footnote{Note that for the NRG models, we only use responses  produced within a pre-defined timeout period.}, and asked them whether each response candidate is appropriate given the certain dialog context. An annotator can label multiple responses or none of them as appropriate. To determine if a response is appropriate we ask annotators to see if the response is relevant to the dialog context and that it does not contradict what was said in previous dialog system's responses. For data annotation we randomly sampled a subset of RUI that contain dialogs with more than 5 turns and fewer than 30 turns. A snapshot of the interface for the annotation task can be found in Appendix~\ref{annotation_details}.

We randomly split the annotated conversations into training and test sets. 
Table \ref{dataset_statistics} shows the statistics of our annotated response selection data, denoted as RSD. Due to user privacy constraints, we cannot release this data. Note that we assume our response selector must always choose a response and therefore we drop turns where none of the responses are labeled as appropriate, and for each turn, we may have multiple positive and negative responses. 

\subsection{RSD Training Variations} 
\label{subsec:data-variant}

To show the importance of using hard negative and multiple positive candidates for response selection, we have also created five variations of the train set of RSD. In our experiments, for each variation we ran random sampling five times and report the average results. 

\begin{itemize}

    \item RSD Train with one positive candidate (denoted as ``RSD 1 Pos.''). 
    Based on the original RSD Train, we sample only one positive candidate for each turn from the multiple positive candidates, and keep all the annotated negative responses. This leads to 8,046 positive and 78,273 negative candidates.

    \item Synthetic Inter-Random. 
    Based on the above-mentioned ``RSD 1 Pos.'' set, we further remove the human annotated negative candidates, and instead use five randomly selected responses from other dialogs and deem these as the new negative candidates. There are 8,046 positive and 40,230 negative candidates in this set. This approach to constructing negative candidates is commonly used in the literature. We experimented with different number of negative candidates and found sampling 5 negative candidates at each turn had the best results.
    
    \item Synthetic Intra-Random.
    Similar to the above set, we use one positive example and four randomly selected responses as negative, two drawn from a random different dialog and the other two from the same dialog as the candidate we are training on. This set contains 8,046 positive and 32,184 negative candidates. This approach to constructing negative candidates is proposed by~\cite{han-etal-2021-fine}. We experimented with different number of negative candidates and found sampling 4 negative candidates at each turn had the best results.
    
    \item Synthetic Adversarial.
    Based on the above-mentioned ``RSD 1 Pos.'' set, we further create negative candidates using the Mask-and-Fill approach from \cite{gupta2021synthesizing}. This approach uses the hierarchical masking function from~\cite{donahue2020enabling} to replace spans in a positive example with blank tokens that will be replaced with tokens predicted from an Infilling Language Model from ~\cite{donahue2020enabling}. For every turn, an average of 28.22 negative candidates were constructed using this approach. We experimented with different number of negative candidates and found sampling 10 negative candidates at each turn had the best results. In total, we have 8,046 positive and 76,307 negative candidates. 
    
    \item Synthetic Retrieval.
    In this approach, we generate negative examples that are semantically similar to the positive example. This approach to constructing negative candidates is proposed by~\cite{li-etal-2019-sampling}. The motivation behind this approach is to create negative candidates that are somewhat similar to the positive candidate and use these as hard examples for the model to train on.
    Specifically, we use the ~\textit{all-MiniLM-L6-v2} model~\cite{wang2020minilm} from HuggingFace\footnote{https://huggingface.co/sentence-transformers/all-MiniLM-L6-v2} and create a sentence embedding for each response in our dataset. At each turn we compute the cosine similarity between the positive candidate and all the other responses in the dataset. We then take responses that have a cosine similarity between 0.8 and 0.95 as a negative candidate. We experimented with different thresholds and found this had the best results. Using these thresholds we get an average of 2.2 negative candidates per turn.  In total we have 8,046 positive and 17,778 negative candidates. 
    
\end{itemize}

\section{Response Selection Models}

We have adopted two state-of-the-art methods for response selection and adapted them to our new dataset for a comprehensive empirical evaluation.

\subsection{DialogRPT~\cite{gao2020dialogue}\footnote{https://github.com/golsun/DialogRPT/}}

DialogRPT is initialized with DialoGPT~\cite{zhang2019dialogpt} and trained using a contrastive loss function to predict a higher score for the positive response given the dialog context and a pair of one positive and one negative response. 
Trained on the Reddit dataset, five different ranker models are proposed by training DialogRPT on different synthesized labels (see the original paper for details).

\subsection{BERT Models}
We experiment with two different BERT model variants for response selection:

\textbf{BERT-FP} ~\cite{han-etal-2021-fine}\footnote{https://github.com/hanjanghoon/BERT\_FP}: 
BERT-FP has achieved high scores on the Ubuntu Dialogue Corpus test set~\cite{lowe2015ubuntu}. The authors post-train the Masked Language Model (MLM) head and Next Sentence Prediction (NSP) head of a BERT-base model~\citep{devlin2018bert} on the Ubuntu corpus via unsupervised learning. Given a dialog context and a response, the NSP head is trained to predict whether a response is either: the ground truth, from a random dialog, or from a random turn in the same dialog. After post-training, the model is further fine-tuned on downstream data for response selection, where given a dialog context and a system response, the model classifies whether this is the correct response or not. 

~\textbf{BERT-Ranker}: We directly fine-tune a BERT-base~\cite{devlin2018bert} model without the above-mentioned post-training step. We denote this model as BERT-Ranker.

\begin{figure}[!htpb]
 	\centering
 		\includegraphics[width=0.49\textwidth]{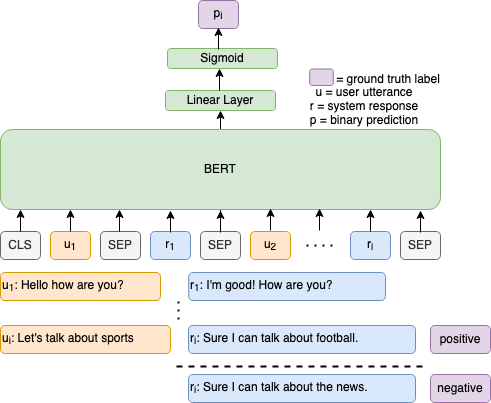}
 	\caption{Model architecture of BERT-Ranker and BERT-FP.}
	\label{fig:response_selection}
\end{figure}

Figure~\ref{fig:response_selection} illustrates the fine-tuning stage for both BERT models. To construct our input, we concatenate the dialog context with a system response and follow the same training procedure used by~\cite{han-etal-2021-fine}, which uses the pooled output representation by the BERT model, passes it through a linear layer followed by a sigmoid function, and minimizes the binary cross-entropy function to predict whether the given system response is positive or negative.

\begin{table*}[t]
\centering
\small
\begin{tabular}{l l c c c c c c}
\toprule
    Model & Train Data & MRR & R@1 & R@2 & R@3 & R@4 & R@5\\ \midrule
     DialogRPT & Reddit & 0.681 & 0.481 & 0.730 & 0.868 & 0.939 & 0.988\\
     \hdashline[.4pt/1pt]
     BERT-FP & Ubuntu & 0.684 & 0.486 & 0.742 & 0.864 & 0.930 & 0.979\\ \hline
     DialogRPT & RSD Train & 0.787 & 0.647 & 0.834 & 0.910 & 0.979 & 0.992\\
     \hdashline[.4pt/1pt]
     BERT-FP & RSD Train & 0.795 & 0.657 & 0.841 & 0.931 & 0.973 & 0.994\\
     \hdashline[.4pt/1pt]
     BERT-R & RSD Train &~\textbf{0.796} &~\textbf{0.659*} &~\textbf{0.843*} &~\textbf{0.936*} &~\textbf{0.980*} &~\textbf{0.995*} \\
     \hdashline[.4pt/1pt]
     BERT-R & RSD 1 Pos. & 0.762(0.06) & 0.628(0.05) & 0.806(0.06) & 0.894(0.07) & 0.941(0.07) & 0.958(0.07) \\
     \hdashline[.4pt/1pt]
     BERT-R & Synthetic IE & 0.688(0.01) & 0.488(0.01) & 0.741(0.00) & 0.880(0.00) & 0.949(0.00) & 0.984(0.00) \\ 
     \hdashline[.4pt/1pt]
     BERT-R & Synthetic IA & 0.698 (0.00) & 0.506 (0.00) & 0.750 (0.00) & 0.879 (0.00) & 0.948 (0.00) & 0.983 (0.00) \\ 
     \hdashline[.4pt/1pt]
     BERT-R & Synthetic Adv & 0.712 (0.00) & 0.532 (0.00) & 0.753 (0.00) & 0.884 (0.00) & 0.950 (0.00) & 0.987 (0.00) \\
     \hdashline[.4pt/1pt]
     BERT-R & Synthetic Ret & 0.718 (0.00) & 0.533 (0.01) & 0.776 (0.00) & 0.902 (0.00) & 0.961 (0.00) & 0.990 (0.00) \\ 
     \bottomrule 
\end{tabular}
\caption{\label{results} Model results on RSD Test. Results for BERT-R (Ranker) using Synthetic datasets are computed by sampling candidates with five different seeds and averaging the model prediction results across those runs. Standard deviations are in parentheses. IE (Inter-Random), IA (Intra-Random), Adv (Adversarial), Ret (Retrieval) are the four different ways of creating negative examples described in Section~\ref{subsec:data-variant}. Recall numbers marked with * mean that the improvement is statistically significant compared with Synthetic Ret (mcnemar with p-value < 0.05).}

\end{table*}

\section{Experiments}
\subsection{Experimental Setup}

Following the previous work~\cite{whang2020effective, han-etal-2021-fine,whang2021response, gu2020speaker, xu2020learning, zhang2021structural}, for evaluation metrics, we use MRR (mean reciprocal rank) and Recall at k (R@k), which is defined as the correct answer existing among the top-k candidates. 

For DialogRPT, we run their five different rankers out of the box over RSD Test in a zero-shot fashion and find that the human vs random ranker scores the highest for both MRR and Recall, therefore we fine-tune this model on RSD Train  following the same training approach in the original paper. Since we have $p$ positive and $n$ negative candidates for each turn, we can obtain $p\times n$ example pairs. For our BERT models, we finetune both BERT-FP and BERT-Ranker on RSD Train. To evaluate the effect of positive and negative examples, we finetune the BERT-Ranker using different RSD training variations described in Section~\ref{subsec:data-variant}.

We also implemented model ensembling for all the methods.
We first divide the training set into five folds, and each time we choose four of them for model training and the remaining one for validation. In this way, we obtain five trained models, and then average their prediction probability outputs on the test set to get the final prediction scores. Further training details are provided in Appendix~\ref{response_selection_details}.

\subsection{Results}

Table \ref{results} shows the results on RSD Test using different models and training configurations. 
From the table, we
have the following findings:
\vspace{-0.05in}

\begin{itemize}
    \item We observe that there is no performance improvement when training BERT-FP on RSD Train versus BERT-Ranker on RSD Train. Therefore the post-training process via optimizing the MLM and NSP objectives proposed in the BERT-FP model does not bring an extra advantage.
    
    \item By comparing DialogRPT trained on both Reddit and RSD Train as well as BERT-FP trained on Ubuntu and RSD Train, we can see that the same models trained on our labeled data lead to much better performance, because of the matched training and testing setup.
    
    \item We observe that training BERT-Ranker on adversarially created negatives (Synthetic Adv.) and (Synthetic Ret.) outperforms using random negatives within the same dialog (Synthetic IA), achieving Recall@1 scores 0.532 and 0.506, respectively. However, training on adversarial examples (Synthetic Adv.) and (Synthetic Ret.) still significantly under-performs training on human-verified hard negatives (RSD Train), which achieved a Recall@1 score of 0.659.

    \item We see the benefit of leveraging multiple positive responses, i.e., BERT-Ranker (RSD Train) outperforms BERT-Ranker (RSD 1 Pos.) with Recall@1 scores of 0.659 and 0.628, respectively. 
    
\end{itemize}

\begin{figure}
 	\centering
 		\includegraphics[width=0.5\textwidth]{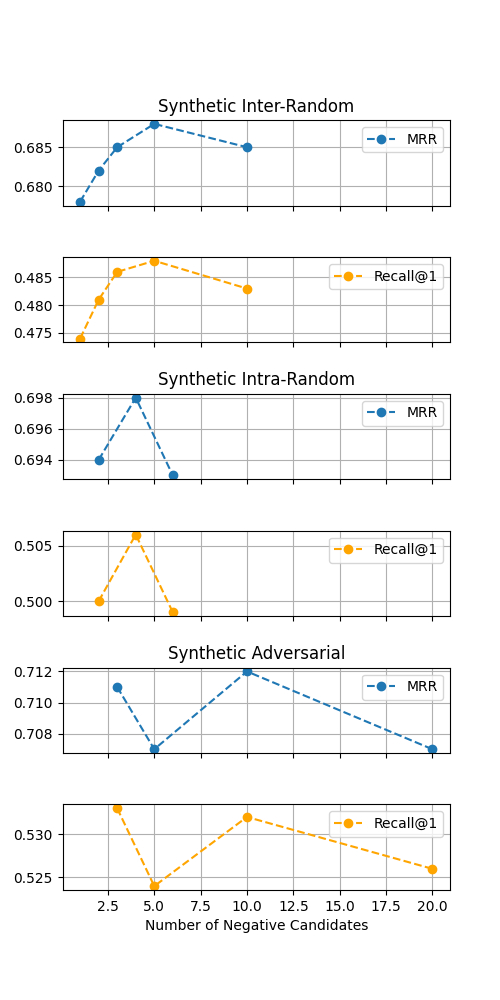}
 	\caption{Results when increasing the number of negative candidates in the Synthetic Datasets Training set. See Recall@2-5 results in Appendix~\ref{evaluation_metrics_for_synthetic}}
	\label{fig:random_negative_abalation}
\end{figure}

\begin{figure}[!htpb]
\small
\centering
\begin{tabular}{p{0.95\linewidth}}
Example 1\\
\hline
{\em Dialog History:}\\
{\bf System:} If you don't mind me asking, what type of music are you into ? \\
{\bf User}: i like rock my favorite band is the arctic monkeys\\
\hdashline[.4pt/1pt]
{\em BERT Ranker (RSD Train):} Thanks for sharing! If you don't mind me asking, what is your favorite rock song ?\\
\hdashline[.4pt/1pt]
{\em BERT Ranker (Synthetic Inter-Random):} i see, you like rock your favorite band is the arctic monkeys. i like rock too. who is your favorite artist?\\
\hline \\
Example 2\\
\hline
{\em Dialog History:}\\
{\bf System:} i'm happy to talk about movies. so, have you seen a good movie lately? \\
{\bf User}: oh yeah\\
\hdashline[.4pt/1pt]
{\em BERT Ranker (RSD Train):} nice. what movie did you see?\\
\hdashline[.4pt/1pt]
{\em BERT Ranker (Synthetic Adversarial):} tom hanks is a great actor. are you a fan?\\
\hline \\
Example 3\\
\hline
{\em Dialog History:}\\
{\bf System:} so do you have any pets at home? \\
{\bf User}: yeah i've got two\\
{\bf System:} cool what is it? \\
{\bf User}: it's a dog\\
\hdashline[.4pt/1pt]
{\em BERT Ranker (RSD Train):} wow! that's great! i like dogs! what is its breed?\\
\hdashline[.4pt/1pt]
{\em BERT Ranker (Synthetic Retrieval):} cool! what kind of pet?\\
\hline \\
\end{tabular}
\caption{Example predictions of BERT Ranker (RSD Train) outperforming BERT Ranker trained on Synthetic Datasets. Due to privacy concerns, these example dialogs are from an internal author.}
\label{fig:bert_past_examples_good}
\end{figure}

\begin{figure}[t]
\small
\centering
\begin{tabular}{p{0.95\linewidth}}
Example 1\\
\hline
{\em Dialog History:}\\
{\bf System:} How about we talk about movies, news or music. So, what topic would you like to talk next ? \\
{\bf User}: tell me about\\
\hdashline[.4pt/1pt]
{\em Model selected response:} ok. i can fetch ratings from rotten tomatoes, check for fresh movies in the theatres, tell you movie plot and reviews, and more.\\
\hdashline[.4pt/1pt]
{\em A correct response:} i didnt hear you clearly, could you please repeat?\\
\hdashline[.4pt/1pt]
{\em Issue:} User utterance 'tell me about' was incomplete due to ASR error. The model did not recognize the error and focused on the dialog topic instead which was movies. \\
\hline \\
Example 2\\
\hline
{\em Dialog History:}\\
{\bf System:} so, tell me, what was the last movie you liked? \\
{\bf User}: i don't remember\\
\hdashline[.4pt/1pt]
{\em Model selected response:} i don't know this movie. space odyssey 2001 is great, in my opinion. it is a long-standing classic, one of the first films about artificial intelligence. tell me, what movie genres do you prefer? \\
\hdashline[.4pt/1pt]
{\em A correct response:} i have found some popular new movies: uncut gems and the rest of us. do you want to know details about uncut gems?\\
\hdashline[.4pt/1pt]
{\em Issue:} The model assumes the user fulfilled the system's question by providing a movie even though the user didn't.\\
\hline
\end{tabular}
\caption{Example predictions of BERT Ranker (RSD Train).
Due to privacy concerns, these example dialogs are from an internal author.}
\label{fig:bert_past_examples}
\end{figure}

\subsection{Analysis}

The advantage of creating negative examples via random or synthetic approaches is the ability to automatically increase the number of training examples. To further evaluate this, we vary the number of negative candidates in Synthetic Inter-Random, Intra-Random, and Adversarial, and report the corresponding MRR and Recall@1 scores, in Figure~\ref{fig:random_negative_abalation}. We see that for our Synthetic Datasets increasing the number of negative candidates to a certain point improves performance for both MRR and Recall@1, after which the performance will degrade.

Increasing the number of negative candidates for (Synthetic Inter-Random) and (Synthetic Intra-Random) increases the likelihood of retrieving a candidate that is a false negative. This can bring noise and confusion to the model during training time. Increasing the size of the corpus could mitigate this issue; however, it can be expensive to collect a large enough dataset to see its benefits.\footnote{Large datasets such as Reddit are known to be noisy and could degrade performance.} The advantage of (Synthetic Adv.) is the ability to create a large number of negative candidates without collecting more data; however, as seen in Figure~\ref{fig:random_negative_abalation} the decrease in MRR and Recall@1 when sampling more candidates may be due to false negatives and therefore still need to be manually verified.

\subsection{Qualitative Examples}

We provide examples of our BERT-Ranker models in Figure~\ref{fig:bert_past_examples_good}. In Example 1 both responses selected by the models acknowledge the user's artist preference; however, BERT-Ranker (Synthetic Inter-Random) chooses a response that repeats the question already answered by the user while BERT-Ranker (RSD Train) does not. In Example 2, BERT-Ranker (RSD Train) provides a more coherent response versus BERT-Ranker (Synthetic Adversarial) which has an abrupt topic change. In Example 3, BERT Ranker (Synthetic Retrieval) repeats the same question asked in the dialog history.

Figure~\ref{fig:bert_past_examples} shows two typical erroneous examples. In the examples we also provide an explanation for the errors.
It is worth pointing out that incorrect ASR output (word errors or end point detection errors such as the first example) is a source of errors to confuse our models.~\citet{gopalakrishnan2020neural} has observed similar issues for the task of response generation in speech-based dialog systems. Future work such as training on synthetic/actual ASR errors is needed to improve the robustness of models for such ASR issues.

\subsection{Limitations}
Our evaluation is done on a dialog dataset that contains a limited number of responders and only GPT2 is used as a neural response generation model. Synthetically created examples may perform better on datasets with a wider variety of neural response generation models. Future work would involve collecting response selection data annotated with a wider variety of responders.

\section{Conclusion}

In this work, we have curated a new dataset for response selection, which contains multiple positive responses and human verified hard negatives. We conducted a comprehensive evaluation of SOTA response selection models and various techniques to construct negative candidates to demonstrate the benefit of the dataset. Even though RSD requires manual annotation we see that training on our dataset greatly outperforms methods that use only one positive example and generate adversarial negative candidates.

\section{Ethics and Broader Impact}
Our work involves re-ranking responses from a dialog system. We acknowledge that we are using data from real users who have not been paid for these interactions. We also acknowledge there may be biases in the demographics of the user population.

\bibliography{anthology,rs2022}
\bibliographystyle{acl_natbib}

\newpage
\clearpage
\appendix

\setcounter{table}{0}
\setcounter{figure}{0}
\setcounter{footnote}{0}
\renewcommand\thetable{\Alph{section}.\arabic{table}}
\renewcommand\thefigure{\Alph{section}.\arabic{figure}}

\label{sec:appendix}

\section{Response Selection Model Training Details}
\label{response_selection_details}
All our BERT-base~\cite{devlin2018bert} models are trained with a batch size of 32 on 1 NVIDIA V100 GPU with 16GB memory. We use the Adam optimizer with a learning rate of 1e-5 and the model is trained for 2 epochs. We use a sequence length of 256 tokens. To deal with the label imbalance, we compute a weighted loss where the loss for a positive candidate is up-weighted by a factor of $\alpha$ and the loss for a negative candidate is down-weighted by a factor of $\beta$. We follow~\cite{king2001logistic} and compute $\alpha$ by taking the sum of the number of positive and negative candidates and divide by the number of labels times the number of positive candidates. The same is done for $\beta$ but we divide by the number of negative candidates instead. In our experiments $\alpha$ = 5.35 and $\beta$ = 0.55.

For the DialogRPT-human vs ranker model, we train with a batch size of 4 on 8 NVIDIA V100 GPUs with 16GB memory each. We use the Adam optimizer with a learning rate of 3e-5. We use a sequence length of 50 tokens and the model is trained for 3 epochs.

\section{NRG Training Details}
\label{nrg_training_details}
We train all our NRG models on the RUI dataset described in Section~\ref{sec:data}. This dataset is split into a 90/10/10 train, valid, test split. All of our models are initialized with GPT2~\cite{radford2019language} based models and were trained with a batch size of 2 on 8 NVIDIA A100 GPUS with 32GB memory each. We use the Adam optimizer and a learning rate of 6.25e-5. Each model is trained for 3 epochs and we finetune both the Language Modeling Head and Multiple Choice Head of GPT2 in a TransferTransfo fashion~\cite{wolf2019transfertransfo}. The Multiple Choice Head is finetuned with 1 randomly selected negative candidate. We leverage the HuggingFace's transformers library for all our models.\footnote{https://github.com/huggingface/transformers} Detailed descriptions of our NRG variants are provided as below.

~\textbf{~\textit{NRG Responder}}: Is a GPT2-XL model where the input is the dialog context which is truncated to 64 tokens.

~\textbf{~\textit{NRG Responder GPT2-medium}}:  Is a GPT2-medium model where the input is the dialog context which is truncated to 64 tokens.

~\textbf{~\textit{NRG Responder grounded on knowledge}}: Is a GPT2-XL model where the dialog context is truncated to 256 tokens and a single knowledge sentence is truncated to 32 tokens. The dialog context and knowledge sentence are concatenated together to be used as input into the model.

~\textbf{~\textit{NRG Responder grounded on dialog acts (DA)}}:  Is a GPT2-XL model where the dialog context is truncated to 64 tokens and each dialog act has it's own embedding that is randomly initialized and updated during finetuning. The dialog context and DA are concatenated together to be used as input into the model. When training this model we automatically label the RUI dataset with a dialog act tagger~\footnote{We annotate a subset of the RUI dataset for dialog acts and train an RNN model on these annotations} and use those DAs as the ground truth. The DA labels used are from~\cite{mezza2018iso} e.g. Feedback, Yes-No question, Statement.

During inference, a sequence of dialog acts are determined using a rule-based dialog policy which are used as input into the model to control the generated response. For example, a Yes-No question dialog act will cause the model response to generate a question~\cite{hedayatnia2020policy}.

~\textbf{~\textit{NRG Responder grounded on sentiment}}: Is a GPT2-XL model where the dialog context is truncated to 64 tokens. There is an embedding representing negative sentiment that is randomly initialized and updated during finetuning. The dialog context and negative sentiment are concatenated together to be used as input into the model to control the generated response. This controllability allows the model is able to generate a sympathetic response when the user expresses negative sentiment. When training such a model, we automatically label the RUI dataset with an off the shelf sentiment classifier~\cite{zhou2020condolences} and use those sentiment tags as the ground truth.

\section{Response Selection Annotation Details}
\label{annotation_details}
Our annotation framework is shown in Figure~\ref{fig:annotation}. A human annotator is shown a dialog context and a set of response candidates are shown below. The annotator can then check off however many responses they deem as appropriate with respect to the dialog context. All responses not selected are considered inappropriate.
\begin{figure}[h]
 	\centering
 	\includegraphics[width=\linewidth]{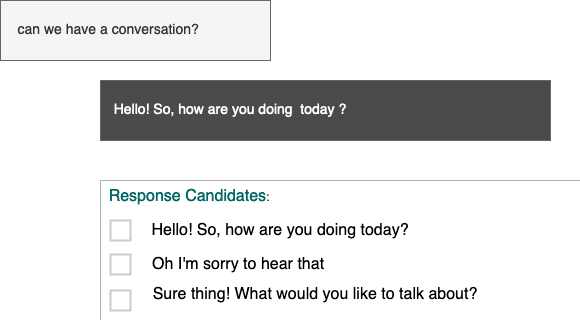}
 	\caption{Annotation framework to collect RSD Train/Test. Due to privacy concerns, this example dialog is from an internal author.}
	\label{fig:annotation}
\end{figure}

\section{Evaluation Metrics for Synthetic Datasets}
\label{evaluation_metrics_for_synthetic}
\begin{figure}[h]
 	\centering
 	\includegraphics[width=\linewidth]{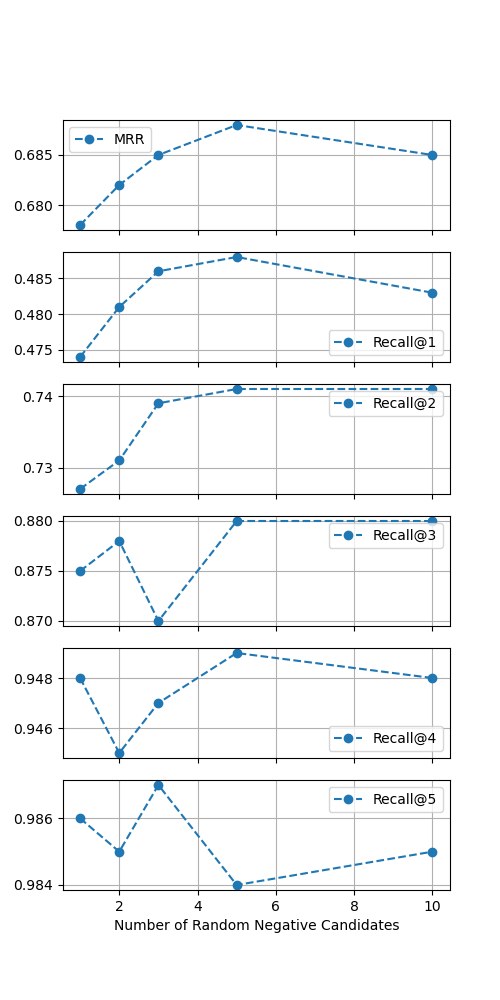}
 	\caption{Increasing the number of randomly sampled negative candidates in the Synthetic Inter-Random Training set.}
	\label{fig:random_negative_abalation_all}
\end{figure}

\begin{figure}[h]
 	\centering
 	\includegraphics[width=\linewidth]{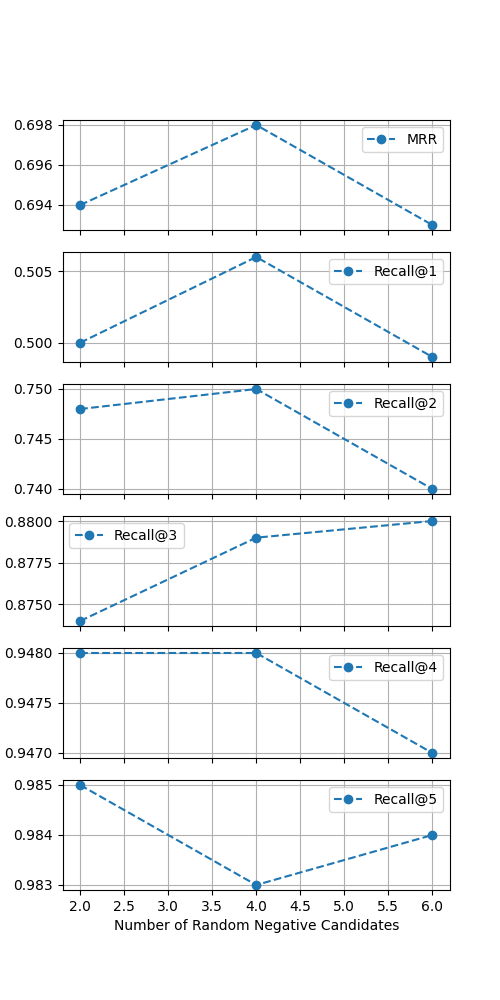}
 	\caption{Increasing the number of randomly sampled negative candidates in the Synthetic Intra-Random Training set.}
	\label{fig:intra_random_negative_abalation_all}
\end{figure}

\begin{figure}[h]
 	\centering
 	\includegraphics[width=\linewidth]{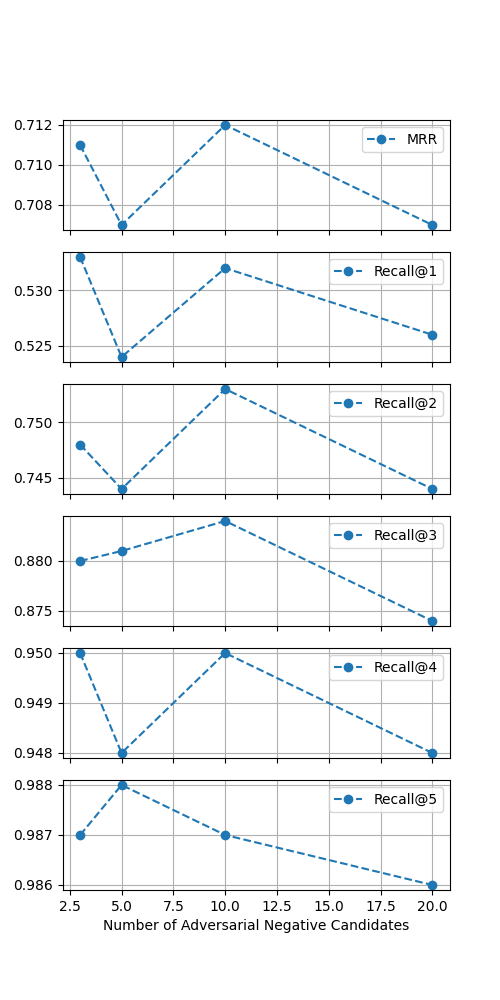}
 	\caption{Increasing the number of randomly sampled negative candidates in the Synthetic Adversarial Training set.}
	\label{fig:adversial_negative_abalation_all}
\end{figure}

In Figures~\ref{fig:random_negative_abalation_all}, ~\ref{fig:intra_random_negative_abalation_all} and ~\ref{fig:adversial_negative_abalation_all} we show all the metrics for our BERT Ranker model trained on each of our Synthetic Datasets with different number of sampled negative candidates. We see for all metrics as the number of negative candidates increase results either degrade or taper off.

\end{document}